# Not only a lack of right definitions:
# Arguments for a shift in information-processing paradigm

Emanuel Diamant
VIDIA-mant, POB 933, Kiriat Ono 55100, Israel
(emanl.245@gmail.com; www.vidia-mant.info)

*Abstract* – Machine Consciousness and Machine Intelligence are not simply new buzzwords that occupy our imagination. Over the last decades, we witness an unprecedented rise in attempts to create machines with human-like features and capabilities. However, despite widespread sympathy and abundant funding, progress in these enterprises is far from being satisfactory. The reasons for this are twofold: First, the notions of cognition and intelligence (usually borrowed from human behavior studies) are notoriously blurred and ill-defined, and second, the basic concepts underpinning the whole discourse are by themselves either undefined or defined very vaguely. That leads to improper and inadequate research goals determination, which I will illustrate with some examples drawn from recent documents issued by DARPA and the European Commission. On the other hand, I would like to propose some remedies that, I hope, would improve the current state-of-the-art disgrace.

## I. INTRODUCTION

In the last ten years or so, we have witnessed an unprecedented robot expansion into our everyday lives – from the traditional fields of factory floor and industrial applications to much more smart and complex activities in homes, workplaces, hospitals, military and space exploration missions, and so on. In this new and fluid environment, robots are expected to co-habitat with humans, interact with them, closely cooperate, collaborate and communicate with their human masters or team-mates. That, in turn, would require them to possess much more human-like features and capabilities, which are usually designated by umbrella terms such as "consciousness", "cognition", "intelligence". This challenge for robotic intelligence has been readily met by the international research community, and since 2002 DARPA and the European Commission have launched a number of research initiatives generally labeled as "Cognitive Robotics", "Machine Consciousness", "Machine Learning", and analogous buzz-word titles. However, generous funding and general sympathy do not ensure success and work progress. Unfortunately, the results of these ambitious research entertainments are far from being satisfactory. The reason for this is twofold: First, the notions of cognition and intelligence, which are usually borrowed from human behavior studies, are notoriously blurred and ill-defined. And second, the very basic concepts underpinning the whole discourse are also either undefined or are very vaguely defined. It is clear and it goes without saying: you could not place the cart before the horses. You could not talk about cognition and intelligence prior to deciphering what you mean regarding the notions of "data", "information", and "knowledge".

## II. DATA, INFORMATION, KNOWLEDGE

The current state-of-the-art is pretty well represented by two recent publications: C. Zins, "Conceptual Approaches for Defining Data, Information, and Knowledge" [1], and S. Legg & M. Hutter, "Universal Intelligence: A Definition of Machine Intelligence" [2]. Both have appeared in the same 2007 year. Both are an extended list of definitions contributed by leading scholars in the fields. The collections reflect systematic and comprehensive thinking and are based on solid theoretical and philosophical foundations. In [1] 45 scholars share their thoughts about Data-Information-Knowledge issues and the article documents 130 definitions of the denoted subjects. In [2] the authors are reviewing 70+ definitions of intelligence. What these two collections undoubtedly exhibit, what these two collections have in common is that definitions offered by the leading scholars in each field have nothing in common among them, and therefore are of little use when it comes to our practical problem-solving. As a result, we are forced to seek for our own definitions. And we are daring to do that.

### A. Data Definition

Data definition usually does not represent any special difficulties. It is simple and intuitive. A compilation of definitions borrowed from [1] would certainly meet our requirements and it looks like this: Data is an agglomeration

(aggregation, collection, assemblage, set) of simple (atomic, elementary) facts (measurements, signs, symbols, characters) resulting from an observation (examination, inspection, monitoring) of the surrounding world. In a human related context data is the output of our senses. In the context of this article data is always digital, that is, it is always represented in a digital form. So, we can finally posit: **Data is an agglomeration of elementary facts**.

B. *Information Definition*

Information definition is the most controversial issue of any discourse. I will not go back to the origins of my research on it. I will refer only to my final findings. The source of my inspiration was the Kolmogorov's complexity theory [4]. But I will not bore you with an extended explanation of all its merits and beauties. Only its main and critical points would be reiterated here.

First, the theory views the assemblage of data elements as a "data object" and its concern is: What is the best way to represent a particular data object? The simplest representation is usually a consecutive enumeration of all data elements constituting an object. (A good example of such a case is a digital image. It can be seen as a rectangle of lines filling the image height and a succession of pixels (picture elements) within each line along the image length. What we call a digital image is essentially a data object with a successive arrangement of pixels in a single frame). Kolmogorov's theory seeks a way for the most effective representation for a single data object. Theoretically two extreme cases can be distinguished here: 1) the elements of a data set are absolutely random and 2) they compose observable structures. In the first case the object can be represented only by the original sequence of its data elements. In terms of Kolmogorov's theory, the length of an object description is equal to the length of its constituting elements. In the second case the regularities observed in data structures could be taken into account, thus leading to a more compact and concise description. In terms of Kolmogorov's theory, that results in a compressed version of the original data object and thus its description length is definitely shorter than the original uncompressed object description. This short-length description can be seen as a program which, when being executed, trustworthy reconstructs (reproduces) the original object because it has all the needed information about it. Moreover, it can be implemented as a computer algorithm. For this reason Gregory Chaitin [5] had named it "Algorithmic Information". Essentially, Kolmogorov, Chaitin, and Solomonoff [3], workings separately, were busy with studying the laws of minimizing the length of such a description, which Kolmogorov dubbed as the object's complexity.

What is interesting to us, what we can learn from these studies is that such a short-length compressed description is the information that we are seeking for about a particular data object. (According to this information the data object can be successfully re-constructed, re-established). Another important point that can be learned here: this information is about the structures distinguishable in a digital data object. (In an image, such structures are visible image segments from which an image is comprised of. So, a set of reproducible descriptions of image segments is the information contained in a given image). As a preliminary conclusion, we can posit: **Information is a description.**

One more thing to be learned from the Kolmogorov's theory is that information is an alphabet-based description usually implemented in an appropriated language. That is, to accomplish an information description some language has to be obligatorily used. Indeed, a variety of known languages can be used for this purpose. For example: programming languages, logic description languages, mathematic languages, and of course, our human natural language – the most common, widespread and frequently used language. A byproduct of this assertion is that information description is always reified as a string of characters, a chain of words, in short, a text string. And Kolmogorov's complexity (or the quantity of information in a given description) is literally the length of such a string.

The Kolmogorov's theory prescribes the way in which a data object description has to be created: At first, the most simplified and generalized structure must be described. (Recall the Occam's Razor principle: Among all hypotheses consistent with the observation choose the simplest one that is coherent with the data). Then, as the level of generalization is gradually decreased, more and more fine-grained image details (structures) could be revealed and depicted. This is the second important point (definition), which follows from the theory's pure mathematical considerations: **Information is a hierarchy of decreasing level descriptions** which unfolds in a coarse-to-fine top-down manner.

Relying on such an approach (to an information definition), I have accomplished a series of experiments aimed to exploit the benefits of applying my new theory to image information content extraction and elucidation. The results of these experiments are described in a series of related publications [6], [7], [8]. For space saving, I will not reiterate the content of those publications. Interested readers are invited to visit my website (http://www.vidia-mant.info) where all my publications can be found and freely accessed.

The immediate result of these experiments was the appreciation that successful recovery and description of image structures (successful image segmentation) does not lead to image understanding. The structures that are observed in an image reflect aggregations of nearby data elements on the basis of similarity among their physical attributes (e.g., color or brightness in visual signals, frequency and intensity in audio signals). These "primary" structures, which I prefer to call "physical structures", usually undergo a further grouping and aggregation, which leads to formation of "secondary" structures that can be called "meaningful" or "semantic" structures. (That is what humans usually perceive as meaningful semantic objects, not to muddle with data objects). It seems to be right to call the descriptions of physical structures "physical information" and the description of semantic structures "semantic information". In this regard, we can posit a third important statement: **There are two sorts of information associated with a given data object – physical information and semantic information.**

It is important to emphasize that both physical and semantic information descriptions are similar in that: (1) they are character strings, (2) they are top-down coarse-to-fine evolving hierarchies, and (3) they are implemented in a certain language. There is only a small difference – physical information can be described in a variety of languages while semantic information can be represented only in a human natural language. (The requirement for a language is true for all levels of information processing, however, in the most primitive living organisms semantic information is represented in proto-languages). Therefore the most suitable form of semantic information representation should be a narrative, a story, a tale. The usual top-down evolving structure of such a story (a narrative, a tale) is well known from other linguistic studies. It descents from the story title, abstract, chapter or section partition, via paragraph subdivision and right to separate phrases and sentences which end up with single words (congregations) actually composing a phrase. Further structural descent leads in linguistics to syntaxes. But in our case – the lowest level of a semantic structure is stuffed with physical information which represents the physical structure of a meaningful object designated by the word in a phrase.

I have first described the interrelation between physical and semantic information in my 2008 article [8] and since then it is repeated in all my later publications [17]. Interested readers can find them on my web site.

*C. Knowledge Definition*

Knowledge is also a controversial issue. According to [1] knowledge is often defined as the next, higher level of Data-Information-Knowledge hierarchy, as "information that has been given meaning". Capitalizing on what we have already learned about information we can easily reject such definitions. For us, meaning is always tied with semantics, and we have already defined above what semantic information is. On the other hand, we can easily accept the widespread assumption that knowledge (a human's previous experience) is the basis to which we relate our perception of the surrounding world, judging and comprehending it. However, equipped with the new definition of information we can revise this assumption in the following way: We can posit that input data stream is initially processed (by the human's visual system) and physical information is extracted from it. Then, the extracted physical information is associated with the physical information kept at the lowest level of a semantic information description. If a proper similarity between the two physical information descriptions is attained, climbing up on the semantic hierarchy ladder we can get the semantic label suitable for this physical information, verify the validity of the neighboring information components (the so called context of the particular information description) and after that a whole fragment of a story can be designated to which the input physical information must be related. In such a way the input data can receive its semantic interpretation. That is, the story fragment to which the input physical information can be tied is the meaningful semantic interpretation of this physical information. Therefore we can definitely declare: **Knowledge is semantic information stored in the system's memory**.

A very important assertion must be emphasized here: Knowledge is not the higher level information, knowledge is memorized semantic information reified in a natural language story, a narrative, a tale.

### III. The Consequences of the Redefinitions

Now, as the basic elements of our discourse have reached their proper resolution, we can start to tackle the issue of our prime concern – the definition of cognition (intelligence) and its possible reification in an artificial design.

*A. Intellegence Definition*

At the first glance, our definition of intelligence (cognition) does not differ from other definitions which are commonly used in Artificial Intelligence. They rely on a following line of arguments: Intelligence is a characteristic feature of a brain, brain is an information processing system, and therefore intelligence is a product of information processing. Google's inquiry for a phrase "Brain is an information processing system" returns 6520 successful hits – a good enough confirmation of the validity and adequacy of such an expression. However, the uncertainty that exists in definitions of Data-Information concepts usually leads to critical subject bewilderment – people talk about information but actually have in mind data that is somehow associated with what they think is information. In the previous sections we have resolutely dissociated these notions. Equipped with our new understanding of Data-Information-Knowledge relationships, we can posit decisively: **Intelligence (cognition) is the system's ability to process information. (Where by information we mean semantic information reified as a natural language story)**.

This is the most general definition of intelligence that we can propose. It fits well all types of cognitive systems: natural (biological) and man-made (artificial) systems, starting from bacteria, plants, animals and human beings, and finally embracing robots and intelligent machines. It coincides well with another recent definition derived from a discussion held within a framework of the Cognitive Systems Foresight Project (commissioned by the Office of Science and Technology, UK government). Their definition states: "Cognitive systems are natural or artificial information processing systems, including those responsible for perception, learning, reasoning, decision making, communication and action", [9].

However, since our notion of "information" strongly deviates from the sense in which other people commonly use it, I presume that there is a need of some more explanations about how the term "information-processing" must be understood and realized in our case.

As it was explained earlier, our definition of "information" assumes that it (information) has to be seen as a two-part composition of physical and semantic information descriptions. In this regard, information-processing has to be seen as a process in which input physical information is mapped into a suitable natural language story retained in the system's memory, where the story provides the frame for the physical information interpretation. (As you remember, according to my theory, natural language story is the form in which a semantic information description has to be represented and kept in a system. At the lowest level of a semantic description (hierarchy) a physical information sub-hierarchy is always present and exists. The similarity between this physical information (present at the lowest level of a semantic description) and the physical information extracted from the input signal is the reason for a particular story selection as a target for physical information mapping and interpretation).

To summarize, we can conclude that when speaking about information-processing in a cognitive system we must always have in mind that two forms of information-processing are present in any cognitive act: low-level information processing, which deals with physical information processing, and high-level information processing, which handles the issue of semantic information processing.

*B. Computing with words*

While the essence of low-level information-processing is usually (and at most intuitively) well enough understood and easily caught on (due to the ubiquity of traditional information-processing approaches), the essence of high-level information-processing is quite new and mostly unfamiliar to the wide research community.

The feeling that mainstream low-level information processing does not exhaust the full range of information-processing in cognitive systems was always hovering in the air. In the mid-1990s, Lotfi Zadeh has introduced the "Computing with Words" information-processing paradigm with a clear intention to replace the dominating data-computation-based information-processing standard with a new linguistic-words-based processing paradigm, "inspired by the remarkable human capability to perform a wide variety of physical and mental tasks without any measurements and any computations" [11]. In 2001 Jerry Mendel has further pushed this idea proposing a technique called a Perceptual Computer – an architecture destined for computing with words [13]. However, despite the use of a vibrant expression – Computing with Words – these innovations did not meet the challenge of natural language processing. They have tackled only a very specific problem's aspect: The fuzziness of natural language words. All other aspects of natural language processing have been left untouched.

Our assumption that semantic information processing revolves around a natural language story (processing), or a narrative, or a tale, immediately placed a requirement for skills and methodologies borrowed from the fields that are usually far away from our everyday interests, e.g., computational linguistics, written or spoken speech understanding, natural language texts parsing, etc. Despite my poor familiarity with these fields I am not sure that they have the proper answers which would allow us to meet the challenges that we are faced with. One of these challenges is: The story that is selected to incorporate the sensed physical information is immediately used for further cognitive manipulations – reasoning about the deviations in physical information, consequent decision making, appropriate action planning and execution, and so on. In context of high-level information processing that means: appropriate story deformation, story transformation and even renewed story reproduction on the basis of a story that was just selected for physical information assignment. In other words, we have to cope with tasks unknown and unfamiliar even to computational-linguistics-related communities, tasks such as Natural Language Comprehension, Natural Language Generation, in short – full-fledged Natural Language Processing with all its entangled aspects and issues. I am not sure that we are prepared to such a twist in our Cognitive Robotics framework delineation. But at least we now have a clearer understanding what our research goals should be.

*C.  Data processing trap*

Meanwhile, for reasons just mentioned above, the issue of information-processing in a cognitive system is always perceived as a data processing issue. The situation can even be justified: At first, input and output streams in any cognitive system are definitely data streams; and second, there is a widely accepted metaphor of a computer as a brain's counterpart. That fits well the idea that the main goal of computer-based processing is establishing of a meaningful correspondence between the input-output data streams (perceived as a search for a best suitable algorithm). All this had guaranteed data processing paradigm dominance in all aspects of information-processing. What else a computer can compute? – Only data can be computed, because the computer is a tireless number crunching artisan, and nothing more than that.

Wise people have raised their warning about this rash attitude, but in vain. As already mentioned, Lotfi Zadeh had once pointed out the unique human capability "to perform a wide variety of physical and mental tasks without any measurements and any computations" [11]. About ten years earlier, Aaron Sloman had called attention to a similar point: "Some people think that "information-processing" refers to the manipulation of bit patterns in computers" [10]. And then: "There is much confusion about what "computation" means, what its relation to information is, and whether organisms in general or brains in particular do it or need to do it" [10].

Despite all that was already said above, robots' cognitive capabilities were always tied directly to the might of the computer's processing power – from a million instructions per second (MIPS) at the beginning of the 1990's to about 100 trillion instructions per second (TRIPS) expected somewhere in the nearest future, on the verge of 2020. Hans Moravec predicts that TRIPS processing power will suffice to emulate in robots human-like cognitive capacity [14]. Meanwhile, researchers are working hard to optimize the algorithms that are supposed to capitalize on the available processing power, something about 100 MIPS today.

The disability of the computer-based data-processing approach to resolve the problem of a semantic gap (the impassable disparity between low-level object features and high-level object perception) is another well known issue that substantiates the flaws of data processing paradigm. Vast efforts are devoted in attempts to overcome this gap, but the problem remains unresolved. From my point of view it will never be resolved, since data-processing is not appropriate for semantic information-processing problem solving.

*D.  Semantic data modeling*

The "semantic gap" is only one pitfall that data-processing approach has kept for its adherents. The second one is the issue of "semantic data modeling". People busy with this subject correctly observe that the mass of data elements is usually partitioned (segmented) into natural data clusters, which are called "object features". (As you remember, something similar to this I have called "primary data structures"). They also observe that humans perceive some specific arrangements of these features as meaningful objects. (I call them "secondary data structures"). The usual practice today is to find out the rules and laws that are guiding the aggregation of the disjoint features into a full-blown object. This process is commonly called "semantic data modeling". I have a less pleasing nickname for it.

The reason for my disrespect is as follows: The primary data structures are natural data structures which reflect some similarities among neighboring elements in the data. Therefore, defining them ("describing them" in terms of this article) is certainly a well-grounded procedure that does not raise any objections, because objective (physical) nature laws underpin such a procedure. The aggregation of object's features into a meaningful whole (object) is totally a subjective process. Its rules are established as a convention, a shared agreement between members of a particular group which are involved in a given scene (or an event) watching and observation. Therefore, their shared agreements can not be extracted from the observed primary data structures' interrelations. The knowledge about the rules that underpin secondary structures formation is a property of human observers (or their artificial counterparts) and not an inherent property of the data. In terms of this paper: semantics is not a property of the data; semantics is a property of a human observer. Attempts to extract semantics from data are a fatal misconception stemming from the fallacies of the data-processing paradigm.

### E. *Machine learning delusion*

If semantics can not be extracted from data (because it does not exist there) then attempts to delegate this task to a fast running machine can be regarded as another flop that stems from a blind belief in data-processing-paradigm's supremacy. Indeed, advances in computer processing power are great and remarkable. Indeed, high-speed computer search for patterns in data may be more efficient than human hunting for suitable structures. But only on one condition – secondary structures are really belonging to the data. However, they are not. Secondary structures (and their descriptions accordingly) exist only in the observer's head, in his mind, in his perception, and nowhere else. Therefore, Machine Learning is a long-standing, extensively cherished delusion, originated from the Artificial Intelligence persuasion that machines could be and would be smarter than humans if only we provide them with the required processing power. And since then this conviction is undefeatable.

## IV. SOME CONCLUSIONS

I hope I was fortunate enough to clarify the Data-Information-Knowledge-Intelligence interrelations which are crucially important for Machine Intelligence (Cognitive Robotics) R&D Roadmap delineation. My key points are: Data is an aggregation of elementary facts; Information is a description of structures observable in the data; Two kinds of information must be distinguished – "physical information" and "semantic information", the first deals with primary structures that result from similarities among neighboring elements within the data, while the second is a description of secondary structures which result from primary structures interrelations. Thus, physical information must be seen as an inherent property of the data, while semantic information must be seen as a property of an external observer that watches and scrutinizes the data. Data is not Information, but Knowledge is Information (semantic information memorized in system's memory). And finally, Intelligence is the system's ability to process information, that is, to process word strings and natural language texts which are the most suitable representations of semantic information.

There is no need to emphasize that my definitions strongly deviate from the current mainstream perception on the matters. Denying and rejecting my views, the dominating research authorities (DARPA and European Commission) inevitably slip and fall in their research guidelines determinations. For example, in [15], one of the main goals of Challenge 2 is defined as "the problem of extracting meaning and purpose from bursts of sensor data…" That is a false and a misleading statement – sensor data (as any data in general) does not possess semantic information, and therefore meaning (semantic information) can not be extracted from it. In [16], another confusing statement is hastily declared: "DARPA is interested in new algorithms for learning from unlabeled data in an unsupervised manner to extract emergent symbolic representations from sensory input…" Again, because semantic information is a convention, an agreement, a property shared between a company of particular observers, it can not be learned by any means. It can be exchanged, transferred, relocated between the group members, or between humans and intelligent machines (robots) collaborating with them in a working group, but it can not be learned. That is the point!

You can arrogantly ignore my speculations, but more than sixty years of unsuccessful attempts to reach the goals proudly declared in the field of Artificial Intelligence must oblige you to be more careful at this occasion. Sure, I am not the great Hillel, and obviously have no chance to teach you the whole Bible of Machine Intelligence while you are standing on your one foot. But, at least, I have really tried to do my best.

# REFERENCES

[1] C. Zins, "Conceptual Approaches for Defining Data, Information, and Knowledge", Journal of the American Society for Information Science and Technology, vol. 58, no. 4, pp. 479-493, February 15, 2007.

[2] S. Legg and M. Hutter, "Universal Intelligence: A Definition of Machine Intelligence", Available: http://arxiv.org/abs/0706.3639.

[3] R. J. Solomonoff, "A formal theory of inductive inference", Information and Control, Part 1: vol. 7, no. 1, pp. 1-22, March 1964; Part 2: vol. 7, no. 2, pp. 224-254, June 1964.

[4] A. Kolmogorov, "Three approaches to the quantitative definition of information", Problems of Information and Transmission, vol. 1, no. 1, pp. 1-7, 1965.

[5] G. Chaitin, "On the length of programs for computing finite binary sequences", Journal of the ACM, vol. 13, pp. 547-569, 1966.

[6] E. Diamant, "Searching for image information content, its discovery, extraction, and representation", Journal of Electronic Imaging, vol. 14, issue 1, January-March 2005.

[7] E. Diamant, "Does a plane imitate a bird? Does computer vision have to follow biological paradigms?", In: De Gregorio, M., et al, (Eds.), Brain, Vision, and Artificial Intelligence, First International Symposium Proceedings. LNCS, Vol. 3704, Springer-Verlag, pp. 108-115, 2005. Available: http://www.vidia-mant.info.

[8] E. Diamant, "Unveiling the mystery of visual information processing in human brain", Brain Research, vol. 1225, pp. 171-178, 15 August 2008.

[9] "Cognitive Systems Foresight Project – Highlight Notice", UK government, The Office of Science and Technology, 2003. Available: http://mi.eng.cam.ac.uk/events/cogsci_jun04/reports/highlight_notice.pdf.

[10] A. Sloman, The Computer Revolution in Philosophy: Philosophy Science and Models of Mind, The Harvester Press, Sussex, UK, 1978, last edition 2008, Available: http://www.cs.bham.ac.uk/research/projects/cogaff/crp/crp.pdf.

[11] L. Zadeh, "From computing with numbers to computing with words: from manipulation of measurements to manipulation of perceptions", IEEE Transactions on Circuits and Systems, Part I: Fundamental Theory and Applications, vol. 46, no. 1, pp. 105-119, January 1999.

[12] L. Zadeh, "Toward Human Level Machine Intelligence – Is It Achievable? The Need for a Paradigm Shift", International Journal of Advanced Intelligence, vol. 1, no. 1, pp. 1-26, November 2009.

[13] J. Mendel, "The perceptual computer: An architecture for computing with words", in Proceedings of IEEE International Conference on Fuzzy Systems, FUZZ-IEEE 2001, Melbourne, Australia, December 2001, pp. 35-38.

[14] H. Moravec, Robot: Mere Machine to Transcendent Mind, Oxford University Press, Oxford, New York, 1999.

[15] ICT Work Programme 2009, Challenge 2, Objective 2.1: "Cognitive Systems and Robotics", European Commission Document C(2008) 6827 of Nov. 2008 , Available: ftp://ftp.cordis.europa.eu/pub/fp7/ict/docs/cognition/fp7-wp-2009_en.pdf

[16] DARPA Request for Information (RFI) SN08-42: Deep Learning, Available: http://www.darpa.mil/ipto/solicit/baa/RFI-SN-08-42_PIP.pdf

[17] E. Diamant, "Machine Learning: When and Where the Horses Went Astray?", In: Yagang Zhang (Editor), Machine Learning, In-Teh Publisher, 2010, pp. 1-18, Available: http://sciyo.com/books/show/title/machine-learning .